\documentclass{article}

\PassOptionsToPackage{numbers, compress}{natbib}

     \usepackage[final]{neurips_2020}

\usepackage{color}
\usepackage{subcaption}
\usepackage{graphicx}
\usepackage{algorithmic}
\usepackage{algorithm}

\usepackage{comment}

\usepackage[utf8]{inputenc} 
\usepackage[T1]{fontenc}    
\usepackage{hyperref}       
\usepackage{url}            
\usepackage{booktabs}       
\usepackage{amsfonts}       
\usepackage{nicefrac}       
\usepackage{microtype}      
\usepackage{amsmath}
\usepackage{xspace}

\newcommand{\methodname}[0]{Top-KAST\xspace}
\newcommand{\learningrate}[0]{\eta}

\title{\methodname : Top-K Always Sparse Training}

\author{Siddhant M. Jayakumar \\
DeepMind \\
University College London \\
\And 
Razvan Pascanu \\
DeepMind \\
University College London \\
\And 
Jack W. Rae \\
DeepMind \\
\AND
Simon Osindero \\
DeepMind \\
\And
Erich Elsen \\
DeepMind \\
}

\begin{document}

\maketitle

\begin{abstract}
Sparse neural networks are becoming increasingly important as the field seeks to improve the performance of existing models by scaling them up, while simultaneously trying to reduce power consumption and computational footprint. 
Unfortunately, most existing methods for inducing performant sparse models still entail the instantiation of dense parameters, or dense gradients in the backward-pass, during training.

For very large models this requirement can be prohibitive. 
In this work we propose \methodname, a method that preserves constant sparsity throughout training (in both the forward and backward-passes).
We demonstrate the efficacy of our approach by showing that it performs comparably to or better than previous works when training models on the established ImageNet benchmark, whilst fully maintaining sparsity.
In addition to our ImageNet results, we also demonstrate our approach in the domain of language modeling where the current best performing architectures tend to have tens of billions of parameters and scaling up does not yet seem to have saturated performance. 
Sparse versions of these architectures can be run with significantly fewer resources, making them more widely accessible and applicable. 
Furthermore, in addition to being effective, our approach is straightforward 
and can easily be implemented in a wide range of existing machine learning frameworks with only a few additional lines of code.
We therefore hope that our contribution will help enable the broader community to explore the potential held by massive models, without incurring massive computational cost.
\end{abstract}

\section{Introduction}
The Lottery Ticket Hypothesis~\cite{frankle2018} has spurred interest in training sparse neural networks~\cite{deconstructing_lottery}, as it highlights a prior exciting result -- that only a small subset of weights of a converged model are sufficient to represent the learnt function to high accuracy~\cite{han2015learning, Thimm95evaluatingpruning, exploring-sparsity-rnn, kalchbrenner2018, sparse-connection-1997}. Perhaps even more exciting is the finding of~\citet{kalchbrenner2018} that large sparse models outperform smaller dense models for a fixed parameter and floating point operation (FLOP) budget.

However, while encouraging, the primary method of finding such sparse subsets involves training a \textit{dense} model. While there is a plethora of works proposing increasingly efficient ways to prune dense networks for sparse inference (dense-to-sparse training)~\cite{gupta2018, variational-dropout, Louizos2018}, the field has only more recently begun to look at approaches that start training at the desired sparsity (sparse-to-sparse training) \cite{Mocanu2018, Bellec2017, Mostafa2019, evci2019rigging}.

Additionally, a high performance and scalable sparse-to-sparse approach would considerably benefit the democratisation of deep learning, as state-of-the-art models are ever increasing in size  ~\citep{shoeybi2019megatronlm, kaplan2020scaling, efficientnet}. This increasingly leads to situations wherein state-of-the-art models require large clusters to train which most researchers would have limited access to. The large compute footprints and energy consumption of training such models also  raises important environmental, moral  and economic concerns~\cite{GARCIAMARTIN201975, greenAI, Strubell2019EnergyAP}.

State-of-the-art text-to-speech (TTS)~\cite{kalchbrenner2018, FBwavernnclone} and automatic speech recognition (ASR)~\cite{googleondeviceasr, Pang2018} are other domains that rely heavily on sparsity. Here sparse networks are used for efficient inference on embedded devices as well as to reduce latency. Further, enabling sparse-training could improve models' ability to personalize to different users, and maintain privacy on device ~\cite{edgeprivacy1, edgeprivacy2}.

Sparse training requires both appropriate algorithms and software/hardware to take advantage of sparse operations. Whilst much of the focus in neural network training hardware has centred on accelerating dense linear algebra operations, there is already sparsity support in modern hardware~\cite{nvidiaampere} with more in the development pipeline~\citep{jouppi2017datacenter}.  

Thus, a scalable and performant sparse-to-sparse method promises to unlock large potential benefits to neural network training --- in terms of model scaling, reduced energy consumption and effective inference. The simplest and most scalable of these methods is to simply pick a random static sparse pattern at initialisation and train with this. Approaches such as Sparse Evolutionary Training (SET)~\cite{Mocanu2018} or Dynamic Reparameterization~\cite{Mostafa2019} improve on this by modifying their sparsity masks based on random evolution, but still lag behind corresponding dense-to-sparse methods. More recently, RigL~\cite{Evci2019} is able to match, or supersede the performance of dense-to-sparse methods. It does this by updating sparsity masks by using occasional gradient information. While theoretically entirely sparse, it is difficult to achieve RigL's theoretical bounds and avoid full dense materialization in common deep learning frameworks.
 
In this paper we aim to address some of these issues and propose a fully parameter-sparse training approach called \textbf{\methodname}. Our technique is scalable because it never requires doing a forward pass with dense parameters, nor calculating a dense gradient. It is also easy to implement within existing frameworks.
Briefly, our method consists of selecting a subset of parameters $A \subset \Theta$ that correspond to the top-K parameters by parameter-magnitude for each training step, and applying gradients to a larger parameter subset $B  \subset \Theta$ (where $B \supset A$.)

To avoid the network fixating on a sub-optimal sparse subset, we introduce an auxiliary \textit{exploration} loss to encourage the mask to adapt during training.

We find we are able to get state-of-the-art language modelling performance for small models, when training a Transformer-XL model using~\methodname~on the character-level task: enwik8~\cite{mahoney2011large}. For image modelling,~\methodname~outperforms existing sparse-to-sparse training approaches, such as Sparse Evolutionary Training (SET)~\cite{Mocanu2018} and matches Rigging the Lottery (RigL)~\cite{evci2019rigging} on ImageNet across a range of floating-point operations (FLOPs) budgets.

\section{Method: \methodname}

\begin{figure}
\begin{centering}

\includegraphics[width=0.8\textwidth, trim=0 4cm 0 4cm]{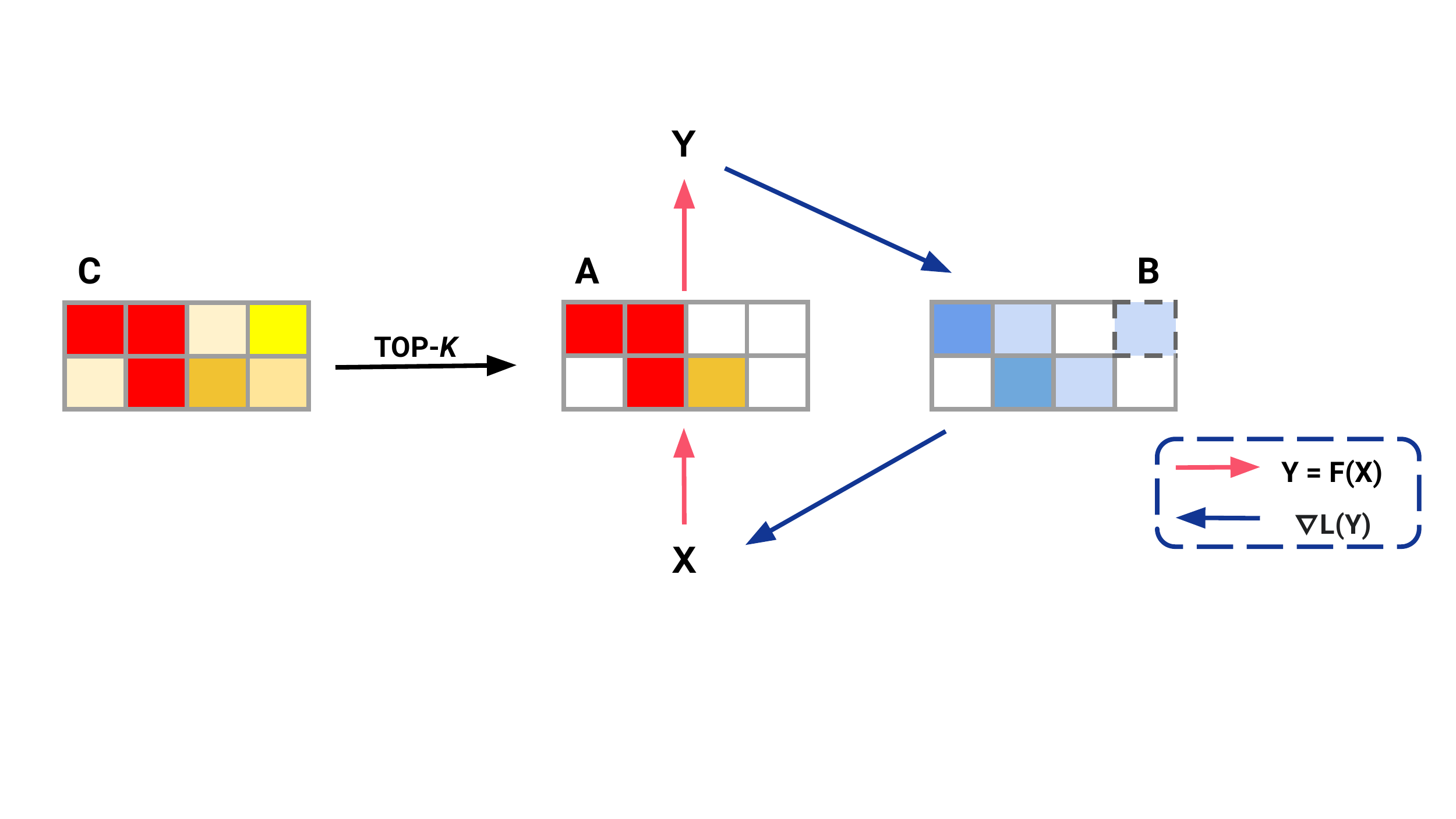}

\end{centering}
\caption{A diagramatic illustration of \methodname. While initialised with an effectively random mask, \methodname explores different permutations by updating an exploration set of weights and choosing the ones with greatest magnitude.}
\label{fig:model}
\end{figure}

\label{desiderata}

The key desiderata for a \textbf{sparse training method}, is that it should:
\begin{enumerate}
    \item Produce a network of desired weight sparsity $S_{final}$ after training is finished.
   
    \item Have minimal compute and memory overheads relative to training a fixed (i.e. static) topology sparse model. 
\end{enumerate}

Dense-to-sparse training methods such as magnitude pruning, Dynamic Neural Wirings (DNW)~\cite{DNW} and Soft Weight Threshold Reparameterization (STR)~\cite{kusupati2020soft} satisfy the first criterion but not the second.  Existing sparse to sparse methods satisfy the second constraint in different ways.  SET and its derivatives occasionally prune unpromising connections and add new ones at random to maintain the same sparsity throughout training.  RigL occasionally prunes unpromising connections and adds new ones based on the locations of the largest gradients from one mini-batch.   We propose an alternate solution that still satisfies the second criterion and achieves high accuracy for a given number of training FLOPs while being easier to integrate into existing frameworks.

\subsection{Sparse Forward Pass}

We consider a generic neural network parameterised by function $f$ with parameters $\theta^t$ at some training step $t$ and input $x$. The output from the forward pass is $y = f(\theta^t, x)$. And during learning the parameters would be updated as $\theta^{t+1} = \theta^t - \learningrate \nabla_{\theta^t} L(y,x)$, where $L$ is the loss function.

Our aim is to to maintain a network weight sparsity of $S \in [0,1]$ throughout training --- where $S$ represents the proportion of weights that are zero ($D=1-S$ is the corresponding \textit{density} proportion of the network).  To do so, at each point in time we consider $\alpha^t$ -- a parameterisation that retains a subset of weights from $\theta^t_i$, and replaces the rest with zeros. We have:
\[
\alpha_i^t = \begin{cases} \theta_i^t & \mathtt{if~}  i \in A^t \\
0 & \mathtt{otherwise} 
\end{cases} 
\]
with $A^t$ used to define a sparse subset of parameter indices that we consider to be ``active'' (i.e. non-zero) at time $t$. 
Membership of $A^t$ is restricted to the top $D$-proportion of weights (from $\theta^t$) by magnitude -- that is: 
\[
A^t=\{i | \theta_i^t \in \mathtt{TopK}(\theta^t, D)\}
\]

In practice, we perform this top-$K$ operation \textit{per layer} instead of on the flattened set of parameters\footnote{Either choice is valid and leads to the same number of parameters.  Global pruning often increases the FLOP requirements by preferring parameters in earlier layers which have more reuse.  It can also suffer from convergence issues at high sparsities due to differing scales in different layers leading to entire layers being pruned.}. One rationale for selecting weights according to their magnitude is that it is an effective but \emph{inexpensive} estimate of which parameters contribute the most to defining the behaviour of the densely-parameterized function $f(\theta, x)$.
Ideally we would like $f(\alpha, x)$ to be the best approximation of $f(\theta, x)$ using $\alpha$ of fixed sparsity-proportion $S$.
To obtain insight into our approximation, we can examine the Taylor series expansion for $f(\alpha, x)$ around $\theta$, where $G$ is the gradient vector and $H$ is the Hessian matrix:
$$f(\alpha, x) \approx f(\theta, x) + G^T(\alpha-\theta) + \frac{1}{2}(\alpha-\theta)^T H (\alpha - \theta) + ...$$
While being able to calculate higher-order derivatives
would provide more accurate sensitivity information~\cite{lecun1990}, it is computationally intractable to do so for very large modern networks. However, as every term in the error scales with powers of $(\alpha-\theta)$, without any information about the higher order derivatives, minimizing the norm of $(\alpha-\theta)$ -- which corresponds to our selection process -- seems the best choice.

During learning we use $\alpha^t$ in both for the forward-pass and in the backward-pass -- hence only incurring the inference and back-propagation compute costs of a sparse model. 
However, $\alpha^t$ is best thought of as a ``temporary view'' of the dense parameterisation, $\theta^t$. That is, the updates will be applied to $\theta$ rather than $\alpha$ and $\alpha^t$ will be reconstructed periodically from $\theta$ by the same deterministic procedure of picking largest (by magnitude) $D$-proportion of weights.

\subsection{Sparse Backward Pass}

The gradient of the loss with respect to a sparse  $\alpha^t$ parameterisation need not result in a sparse gradient vector; indeed the gradient would typically be expected to be fully dense. This is because the gradients with respect to the $0$ entries of $\alpha^t$ need not themselves be zero. This unfortunately would break our key desideratum (2). To avoid evaluating dense gradients we take inspiration from coordinate descent and compute the gradient for a coordinate block composed of parameters with indices from the set $B^t$, where: 
\[
B^t=\{i | \theta_i^t \in \mathtt{TopK}(\theta^t, D+M)\}
\]
By definition, $B$ is a superset of $A$ and contains the indices corresponding to the non-zero entries of $\alpha$ as well as an additional set of indices corresponding to the next largest $M$-proportion of entries (by magnitude) of the dense parameterisation, $\theta$.
Updating the largest $(D+M)$-proportion of weights makes it more likely that this will lead to permutations in the top $D$-proportion weights that are active, and hence allows the learning process to more effectively explore different masks. We refer to this effective sparsity of $(1 - D - M)$ units as our \textit{backward sparsity}. 

Computing the gradient with respect to a subset of coordinates of $\theta$ implies that the gradient we are computing is sparse, and throughout the forward pass and backward pass we do not need to instantiate a dense vector of the size of $\theta$.The final update has the following form\footnote{Our approach is not a strictly valid coordinate descent method on either $\alpha$ or $\theta$.

}:
\[
\Delta_{\theta_i^t} = \begin{cases} -\learningrate \nabla_{\alpha^t} L (y,x, \alpha^t)_i & \mathtt{if~}  i \in B \\
0 & \mathtt{otherwise}
\end{cases} 
\]

At initialisation, $A$ will consist of a random subset of weight-indices from the freshly initialised $\theta^0$. As learning progresses, due to the updates on $B$ coming both from the primary loss and the auxiliary regularisation term (described in detail in the following section) this set will change and evolve the weights and topology most useful for the desired function approximation. %
We postulate learning as going through two stages (and this postulation seems to be observed in practice): 
\begin{itemize}
    \item In the first \emph{exploratory stage}, at each iteration we select a different active set $A$, and its corresponding $\alpha$, and perform one update step on $\theta$ using gradients obtained from the loss on $f(\alpha, x)$ and the regularizer.
    \item In the second \emph{refinement stage}, the active set $A$ effectively becomes fixed, as we settle on a stable pattern of non-zero weights which then undergo fine-tuning to their optimal values. 
\end{itemize}
In the first stage, the updates on the ``additional'' coordinates in the set $B \setminus A$  allows exploration by changing the set of weights that will end up in the active set $A$ (and thus used in $\alpha$) on the next iteration.
In the second stage, these ``additional'' updates will end up being increasingly less impactful and eventually will be effectively ignored, as they will not alter $A$ and hence will not be reflected in $\alpha$ for either the forward or backward passes.
The \emph{exploratory stage} of picking different subsets of parameters from $\theta$ sets makes our approach very different from simply having a fixed random sparsity pattern imposed on the model.

\subsection{Exploration Regularisation Loss}

The method outlined above may lead to a \textit{rich-get-richer} phenomenon: with only the randomly selected weights at initialization being used if others receive insufficient weight updates for their norm to exceed the critical threshold. %
This problem may be particularly pronounced at high levels of sparsity, and to combat it we propose a heuristic inspired by the principle of \emph{optimism in face of uncertainty}, widely used in reinforcement learning (RL)~\cite{BrafmanOptimism}.~%
Concretely, we penalise the magnitude of the weights in set $B$, while those that are neither used nor currently being updated (set $C$) are not penalized at all.
The net effect of this is to reduce the magnitude of the active weights, making it more likely that on the next iteration the algorithm considers new items for the membership of both set $A$ and $B$ — similar to how in RL, optimistic exploration adds bias to favour the selection of actions that have not thus far been chosen often. 

We also posit that for high sparsity settings there is a teetering effect between weights in $B \setminus A$ and $A$ that are very close in magnitude, leading to a slow down in learning. We therefore propose to penalise $B \setminus A$ more than $A$ to increase the critical strength of updates needed for units from $B \setminus A$ to turn on and to stabilise the mask. We heuristically choose the scale to be inversely proportional to $D$, as this effect is more important for $D \ll 1$.

We express this penalty as an $L2$ regularisation, with a similar split of units as above\footnote{The gradient of the regularization term follows the same sparsity pattern as the gradient of the primary loss.}. Specifically: 

\[
Loss_R(\alpha_i^t) = \begin{cases}
|\theta_i^t|  &  \mathtt{if~}  i \in A^t \\
\frac{|\theta_i^t|}{D} & \mathtt{if~}  i \in B^t \setminus A^t \\
0 & else 
\end{cases} 
\]

\subsection{Implementation of \methodname}

As described above,  the compute and memory requirements for \methodname in the forward and backward passes scale with the forward and backward sparsities, respectively. One possible concern is the additional cost of performing a \texttt{Top-K} operation in the forward pass every iteration. While the FLOPs required for this are much fewer than those needed by the actual training --- this could necessitate fitting the dense model in memory. One way to alleviate this is to simply compute the the \texttt{Top-K} entries in parallel on CPU, thus avoiding the need to fit the model on the actual training hardware. The CPU could maintain the parameters in an appropriate data structure, such as a heap that would minimise the cost of updates. Lastly, we show in the sections below that the mask slowly stabilises and in fact we do not even need to perform this operation every step. In appendix \ref{app:impl} we show that we can get comparable results even if we perform this only \textit{every $100$ steps} which significantly reduces communication requirements and extra overheads. 

\section{Related Work}
Methods that require dense weight or gradient information at training time but produce a sparse network at the end of training are now numerous and include: L0 regularization~\cite{Louizos2018}, variational dropout~\cite{variational-dropout}, discovering neural wirings~\cite{DNW}, soft weight threshold reparameterization~\cite{kusupati2020soft}.  Magnitude Pruning is simple and effective~\citep{gale2019state} and we use it throughout as a baseline representative of this class of training methods.  Such methods do not allow us to train larger sparse models than the biggest dense model we could train (in fact it is usually smaller due to overheads).

Sparse training of neural networks first happened through evolutionary means.  Throughout the 1990s there was a flurry a research on the topic of Topology and Weight Evolving Artificial Neural Networks (TWEANNs) exemplified by~\citep{NEAT}.  While the networks \emph{were} sparse during the evolution, this was not the focus of the research and the advantages of the sparseness in terms of enabling size and efficiency were mostly ignored.  There has also been some recent work on using evolutionary methods to evolve sparse topologies~\cite{karelhybrides}.

Deep Rewiring~\cite{Bellec2017} was the first work to consider sparse training of weight-sparse neural networks within the framework of gradient descent.  It restricts weights to have a fixed sign, and sets weights to zero when their sign would flip.  Additionally, it introduces a random walk in parameter space and can be thought of a constrained Monte Carlo sampling procedure over both the weights and the network connectivity.  Despite theoretical convergence proofs, its practical performance seems to lag behind later, less well founded work~\cite{Mostafa2019}.

This was followed by Sparse Evolutionary Training~\cite{Mocanu2018} which uses weight magnitudes to drop weights and introduces new connections at random, drawn from the original initialisation distribution.  It is both simpler and more effective than Deep Rewiring. Our method, \methodname modifies the units based on gradient information instead which we find is more performant than random additions. 

Dynamic Reparameterization~\citep{Mostafa2019} introduces a method for moving a parameter budget between different layers.  This allows the network to better put parameter capacity where it is most effective.  However, this ignores a FLOP constraint - the amount of FLOPs required to evaluate the network can change (usually upwards) because of these modifications.

Lastly, Rigging the Lottery (RigL)~\citep{evci2019rigging} is a recent and highly performant sparse-to-sparse method that matches or surpasses the performance of pruning-based methods. It uses infrequent full gradient calculations to decide which parameters to `wake-up'.  As it only requires knowing the location of the highest values of the gradients, its theoretical cost is proportional to the network sparsity, though this bound is hard to achieve in practice in current DL frameworks. We also compare \methodname to RigL in this paper and find we are able to perform comparably while alleviating the aforementioned implementation issues.

\section{Experiments: ImageNet} 

Our aim in the section below is to demonstrate the efficacy of our method at enabling sparse training of models across different modalities (vision and language), model types (convolutions and attention) and different sparsity regimes. %
We start by demonstrating the efficacy of our method on the ImageNet dataset for image classification, where we train a sparse ResNet-50 as in previous works \cite{evci2019rigging, gale2019state}. This is a commonly used benchmark for sparsity methods, albeit often used in different regimes. We provide full details of model and hyper-parameters in the appendix \ref{app:imagenet}.

\begin{centering}
\begin{figure}[]
\begin{subfigure}{0.32\textwidth}
\includegraphics[width=0.97\linewidth]{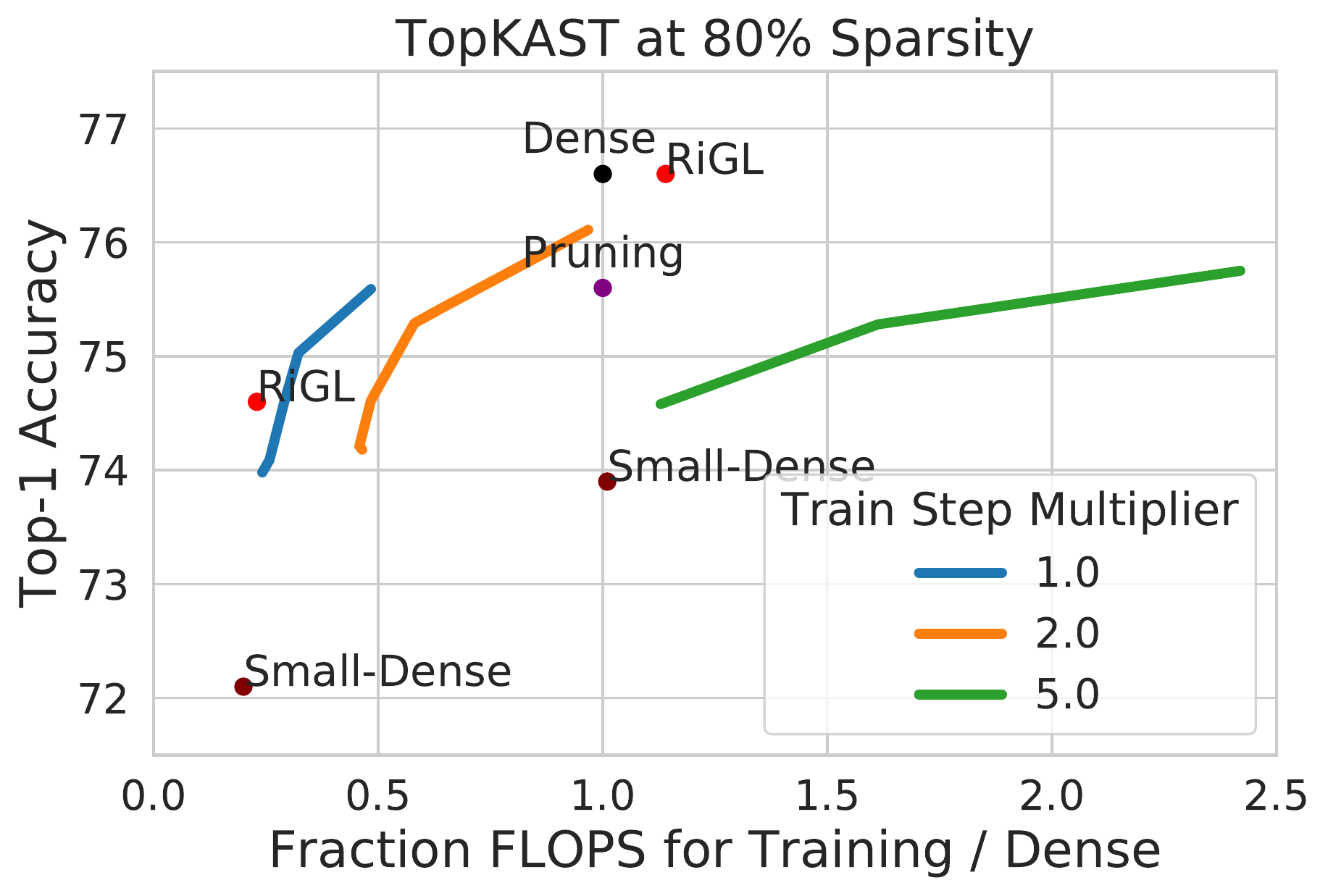}
\end{subfigure}
\begin{subfigure}{0.32\textwidth}

\includegraphics[width=0.97\linewidth]{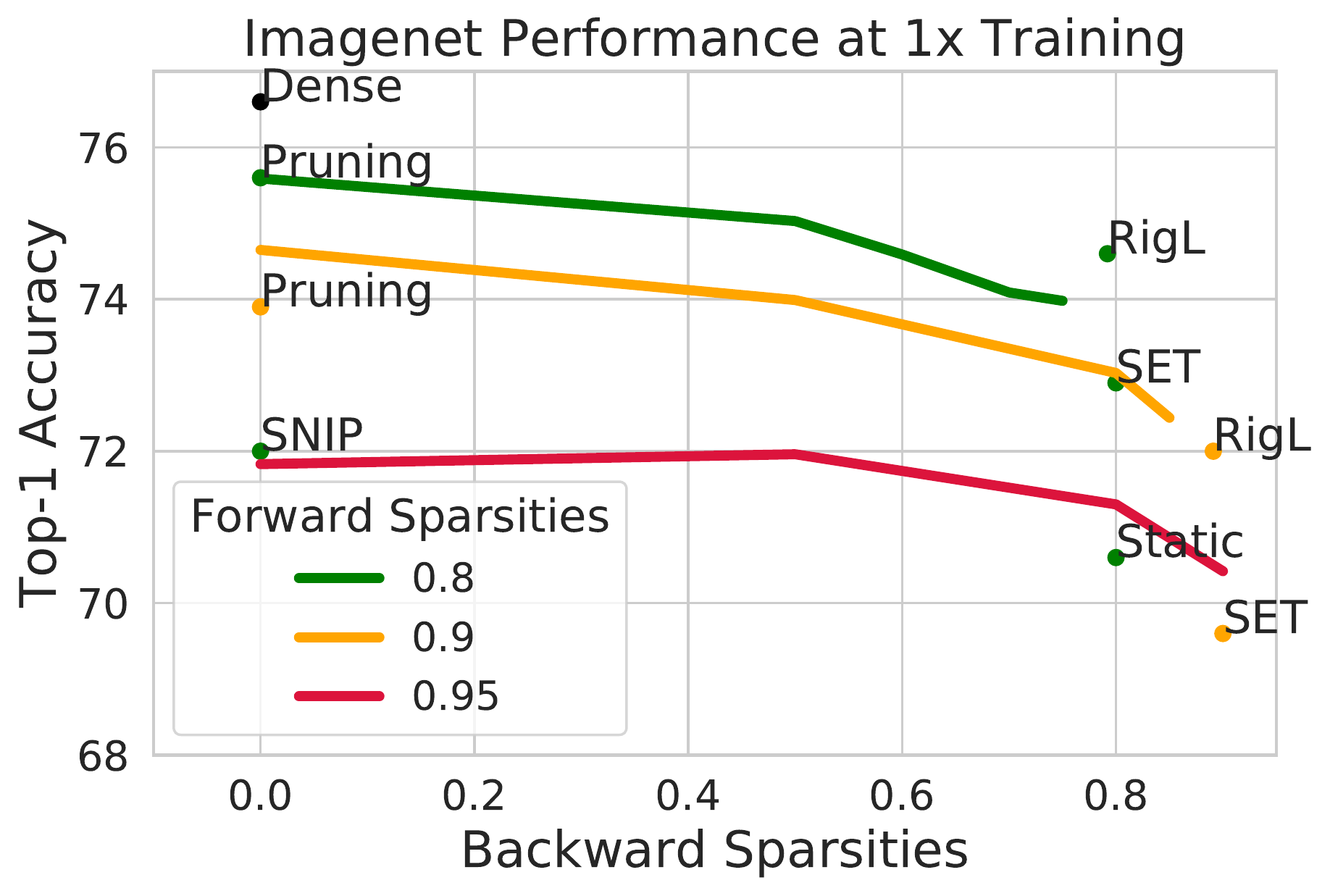}
\end{subfigure}
\begin{subfigure}{0.32\textwidth}

\includegraphics[width=0.97\linewidth]{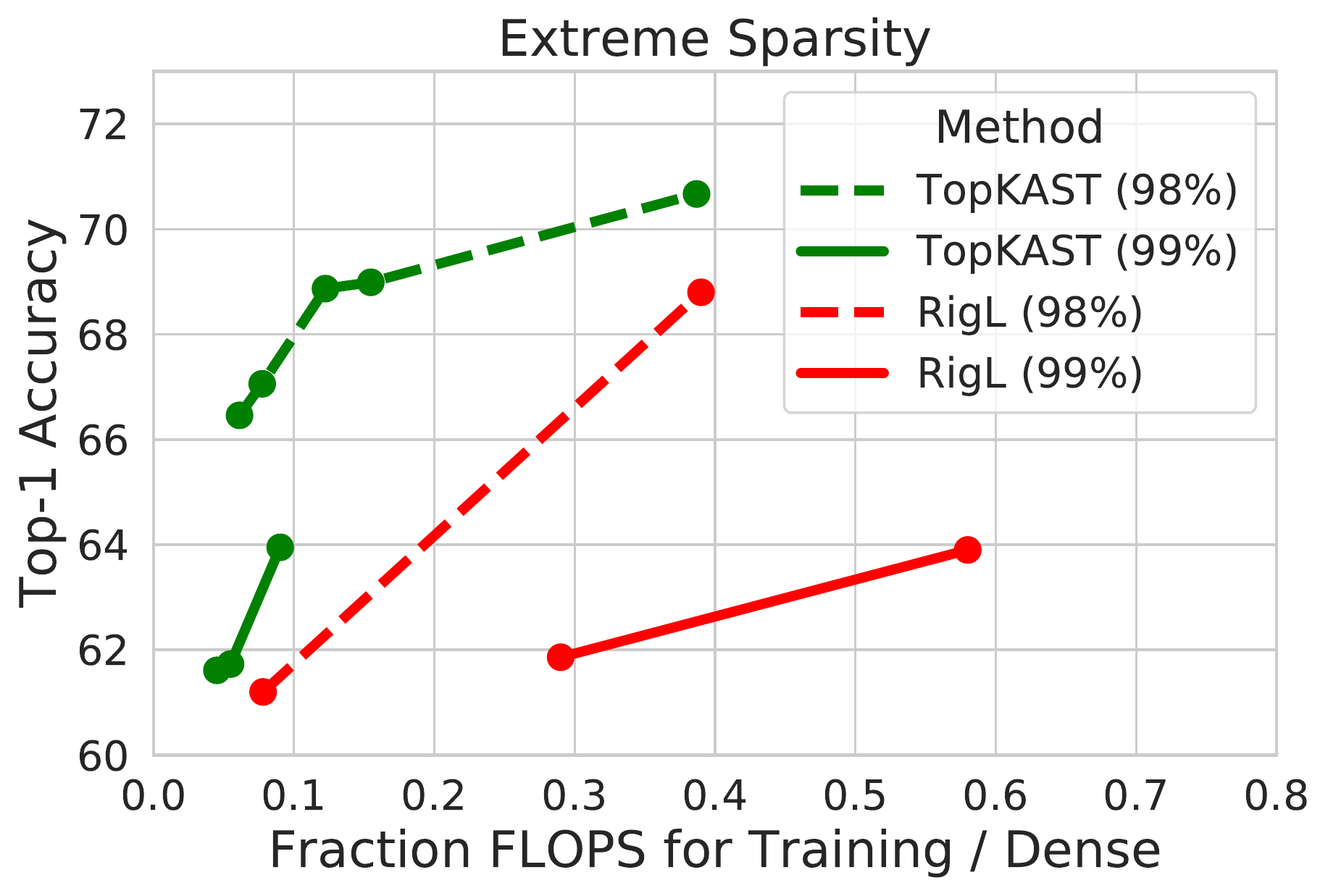}
\end{subfigure}

\caption{(a) FLOPS needed to train various sparse models as a fraction of those for a dense model. The FLOPS for \methodname vary as a function of the backward sparsity and the length of the training run. (b) Comparing methods on the basis of their backward sparsity. (c) \methodname and RigL compared at sparsities of $98\%$ and $99\%$.}
\label{fig:flops80}
\end{figure}
\end{centering}

We first compare methods in the commonly used regime of fixed inference sparsity with first and last layers dense. As \methodname allows practitioners to choose their own level of backward and forward sparsity, we run \methodname for different levels of each, as well for multiples of the default training runs. We summarise this in Figure \ref{fig:flops80} above, showing the spectrum of performance versus FLOPS used (increases with decreasing backward sparsity and increasing training time), for a fixed forward sparsity of $80\%$. We also report results for a variety of standard and state-of-art methods.

We find (Figure \ref{fig:flops80} a and b) that \methodname is comparable (at constant FLOPS) to dense methods like pruning, while advantageously staying completely sparse throughout. \methodname  also outperforms always-sparse methods like SET and Static random sparsity patterns. We further report results for sparsity levels $90\%$ and $95\%$ in \ref{fig:flops80}(b) and results for relaxing the assumption of first and last layers dense, in appendix \ref{app:imagenet}. 

\textbf{Comparing RigL and \methodname}
Fig \ref{fig:flops80} also shows that the most performant prior sparse-to-sparse method is RigL and we see that \methodname performs comparably on a per-FLOP basis. RigL's update of its sparsity pattern  requires  \textit{occasionally} calculating (a top-k over) dense gradients and in Fig 2 (b), we can see that when compared on the basis of average backward sparsity instead, \methodname requires slightly higher densities to match RigL's performance. However, while in theory RigL only needs the \textit{highest} values of this dense gradient, it would require re-writing the gradient calculation for many primitives in existing DL frameworks to achieve this. Additionally, we note that RigL has many hyperparameters that might need tuning: when to start and finish updating the mask, how often to update, the initial drop fraction and the schedule by which this is annealed.  On the other hand, \methodname requires no custom gradient calculations, and the only hyperparameter is the size of bucket $B$, and thus is easier to implement, to use, and is readily scalable. We expand on these implementation details in appendix section \ref{app:impl}. We also find in Fig 2 (c) that \methodname surpasses RigL at higher levels of sparsity ($98\%$ and $99\%$). \methodname's ability to choose slightly higher backward sparsities also means that at the cost of a little extra compute we are able to greatly increase performance.

\subsection{Ablation studies}

\textbf{Selection of $\mathbf{B \setminus A}$.} 

We first consider the question of exploration in the backward pass and the method for selecting set $B$. We defined this set as those units used in the forward $A$ plus the next-highest set of units by magnitude. We can instead consider whether it would not be better to \textit{randomly} sample these extra units. Intuitively we might explore more of the space and in expectation, allow gradient to pass through all units. We see in table \ref{tab:ablations} that this method is far better  for sparsity of $90\%$ but performs far worse for higher levels of sparsity, validating our choice. It is to be expected that this choice becomes more important in very sparse settings, where it would take many iterations to cover relevant weights if they are not directly targeted. Also, randomly picking additional weights means that the mask also changes more through training, whereas we expect the top-$k$ to stay more constant, thus reducing the potential cost of the sampling procedure.

\begin{table}[h!]

\centering
\begin{tabular}{ccccc}
\toprule
    \textbf{Method} & \textbf{Sparsity Forward} & \textbf{Sparsity Backward} & \textbf{Top 1 Acc}  \\ \midrule
     \methodname & $0.9$ & $0.8$ & $73.03$  \\
       \methodname (Random) & $0.9$ & $0.8$ & {$\mathbf{74.76}$} \\
       \midrule
          \methodname & $0.95$ & $0.9$ & \textbf{$\mathbf{70.42}$} \\
   
     \methodname (Random) & $0.95$ & $0.9$ & $68.48$  \\
        \midrule
          \methodname ($t=0$) & $0.9$ & $0.0$ & $68.26$  \\
        \methodname ($t=5000$) & $0.9$ & $0.0$ & $72.05$  \\
          \methodname ($t=16000$) & $0.9$ & $0.0$ & $74.14$\\
              \methodname ($t=32000$) & $0.9$ & $0.0$ & $\mathbf{74.65}$\\
              \bottomrule

\end{tabular}
\vspace{0.5em}
\caption{Ablation Experiments.}
\label{tab:ablations}
\end{table}

\textbf{Analysing the learning dynamics} We can further test our hypothesis that our regularisation, combined with the learning dynamics, divides learning into an \textit{exploration} phase, wherein an optimal mask is discovered, and a \textit{refinement} phase. To do so, we take a standard training run of $32000$ steps and artificially stop the gradient updates to the `extra' units not active in the forward pass ($B \setminus A$). We do so  at different points of training (marked $t$ in Table \ref{tab:ablations}) --- start of training ($t=0$), $t=5000$, or halfway through. We find that removing all exploration units entirely ($t=0$) is very harmful for performance, but training for just $5000$ steps with these considerably boosts performance. At $t=16000$ we have recovered most of the benefits of our method. This provides evidence that for the latter half of training, the gradients fine-tune performance on the learnt mask which stays more or less constant. 

\begin{centering}
\begin{figure}[b]
\centering
\begin{subfigure}{0.32\textwidth}
\includegraphics[width=0.97\linewidth]{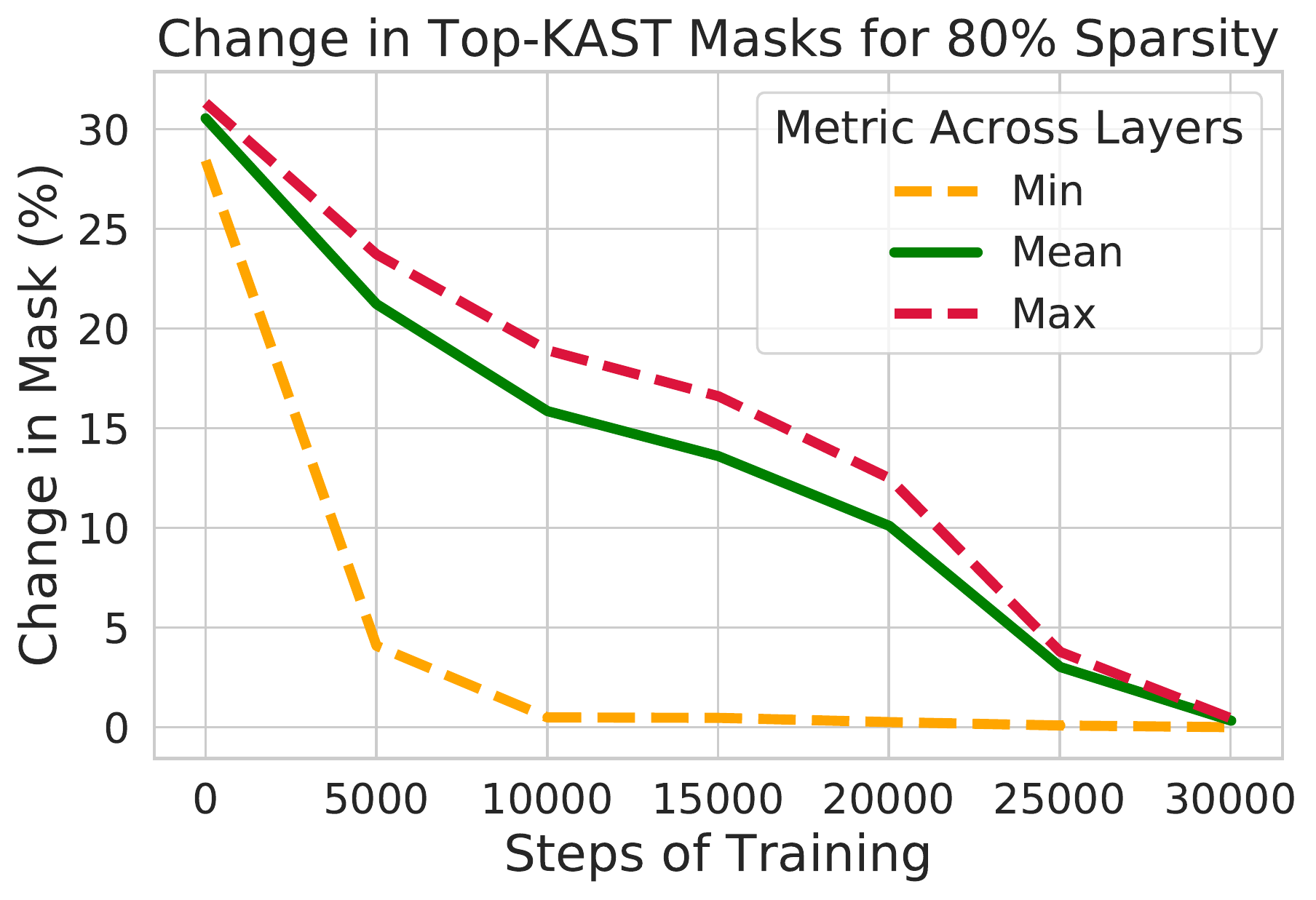}
\end{subfigure}
\begin{subfigure}{0.32\textwidth}

\includegraphics[width=0.97\linewidth]{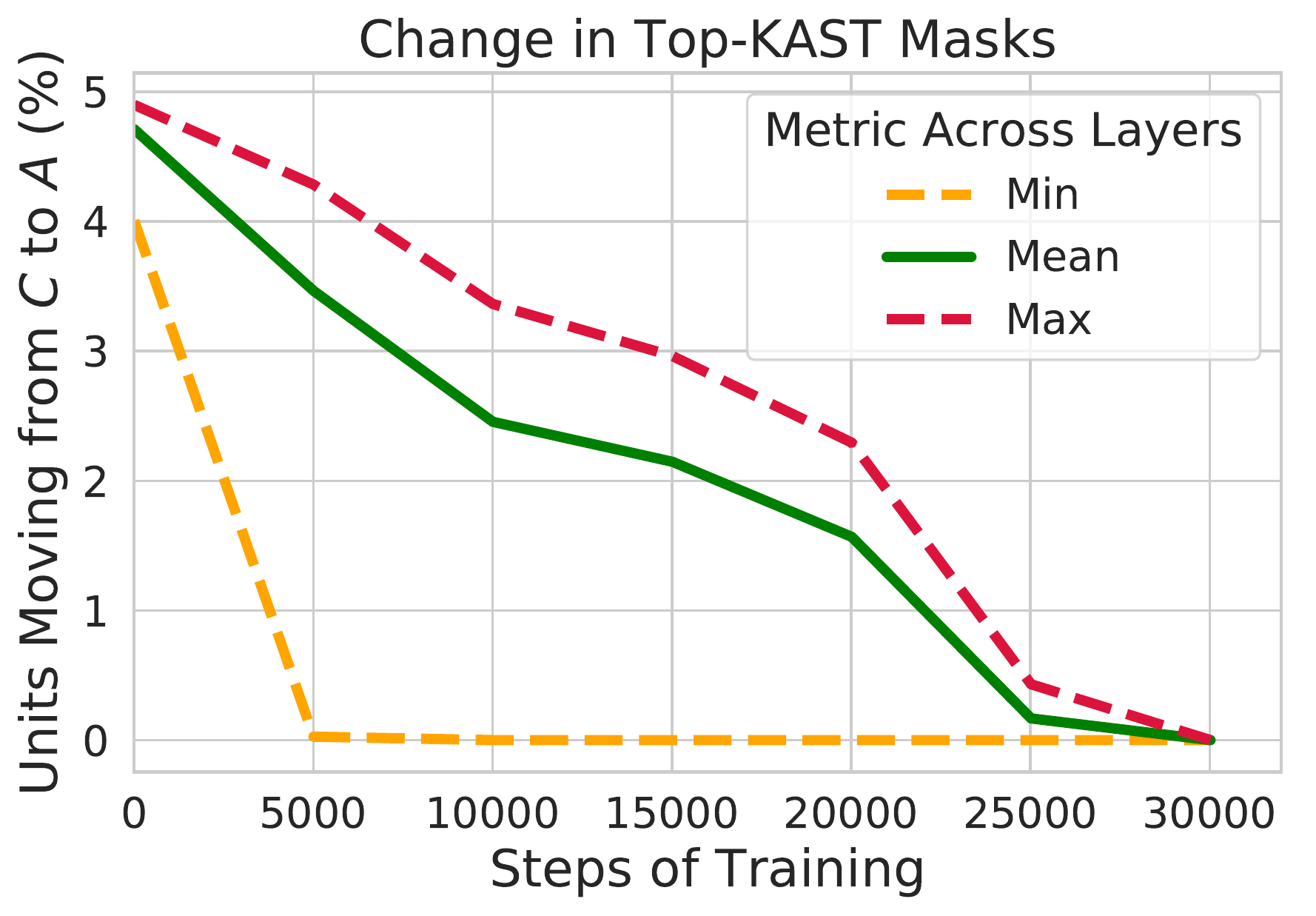}
\end{subfigure}
\begin{subfigure}{0.32\textwidth}
\includegraphics[width=0.97\linewidth]{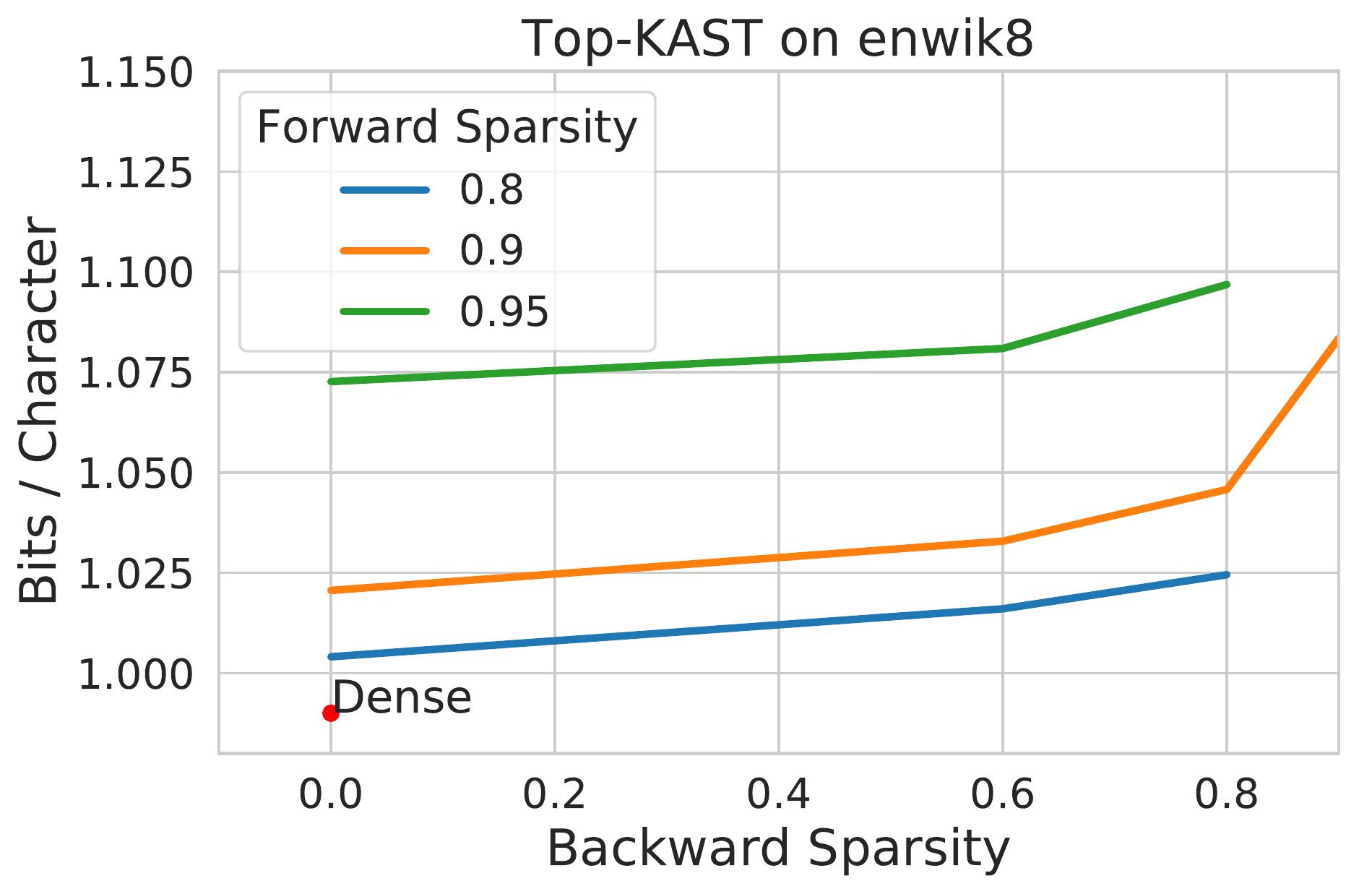}
\label{subfig:enwik8}
\end{subfigure}

\caption{(a) shows that the mask gradually stabilises over time. (b) further, the number of units in set $C$ that make it to the active set $A$ is relatively small and also tends to $0$.} 
\label{fig:measure}
\end{figure}
\end{centering}

\textbf{Analysing the mask dynamics} We can further analyse how the mask changes through time. We take a standard training run as above with forward sparsity of $80\%$ and backward sparsity of $50\%$. We first measure the difference in the sparsity masks $m$ at pairs of points $5,000$ steps apart in training --- i.e. $\frac{(m^t - m^{t+5000}) ^2}{|\theta|} $ --- the fraction of units that change ($m=1$ if the weight is active, else $m=0$). This is summarised in figure \ref{fig:measure} where we show the percentage change in masks across time (we plot min, mean and max across layers). We find that the mask indeed stabilises over time. We can further assess what units that are in set $C$ or the \textit{reservoir} --- units used in neither the forward nor backward passes at initialisation --- ever turn on. We find that only about $5\%$ of these units are ever used and most of this change occurs at the start of training.  This provides more evidence for the exploration and learning dynamics that motivate our design choices. 

\section{Experiments: Language Modeling}
One class of models which have benefited hugely from a greater number of training parameters is language models, notably using the Transformer architecture~\cite{vaswani2017attention, radford2019language}. Language models predict the probability of strings of text, typically by tokenizing the text into a sequence of integers $x_0, \ldots, x_t$ (e.g. characters or words) and then decomposing the joint probability $p(x_0, \ldots, x_t)$ of this sequence into a product of conditional probabilities $p(x_0) \prod_{i=1}^t p(x_i | x_{<i})$.

Language model performance has been observed to follow a power-law of improvement when the data and model parameters are increased~\cite{kaplan2020scaling}. One challenge large parameter sets bring, is an increased strain on memory bandwidth to store the parameters. Approaches which can train and evaluate to comparable performance using less parameters can facilitate eventual training of larger models.  We try~\methodname~to train language models on two commonly-benchmarked datasets: Enwik8~\cite{mahoney2011large} which is a character-level benchmark derived from the \textit{Hutter Prize} and WikiText-103~\cite{merity2016pointer} which is a word-level language model benchmark. We use a long-range Transformer variant, the Transformer-XL~\cite{dai2019transformer}; training hyper-parameters are displayed in Supplementary Section \ref{app:lm}.

\begin{table}[h]
    \centering
    \begin{minipage}[b]{.54\linewidth}
    \begin{tabular}{c c c}
    \toprule
    \textbf{Model} & \textbf{Params} & \textbf{BPC} \\
    \midrule
     Transformer-XL \citep{dai2019transformer} & 277M & 0.99 \\
     \midrule
    Stacked LSTM \cite{graves2013generating} & 21.3M & 1.67 \\
    Hypernetworks \citep{ha2016hypernetworks} & 27M  & 1.34 \\
    mLSTM \citep{krause2016multiplicative} & 46M & 1.24 \\
    Transformer-XL \citep{dai2019transformer} & 44M & 1.06 \\
    All-Attention Transf. \cite{sukhbaatar2019augmenting} & 39M & 1.01 \\
    \midrule
    \textbf{\methodname ($80\%$, $0\%$)} & \textbf{55M} & \textbf{1.00} \\
     \methodname ($80\%$, $80\%$) & 55M & 1.02 \\
      \methodname ($90\%$, $60\%$) & 27.7M & 1.03 \\
    \bottomrule
    \end{tabular}
    \vspace{.5em}
    \caption{\textbf{Enwik8}: test BPC of small models.}
    \label{tab:enwik8_small_models}
    \end{minipage}
    \hspace{1em}
    \begin{minipage}[b]{.40\linewidth}
    \centering
    \begin{tabular}{c c c c}
    \toprule
        \textbf{Fwd} & \textbf{Bwd} & \textbf{Params} & \textbf{Perplexity} \\
    \midrule
         0\% &  0\% & 285M & 18.3 \\
         0\% &  0\% & 94M & 21.5 \\
         \midrule
         80\% & 0\% & 57M & 19.8 \\
         80\% & 60\% & 57M & 21.3 \\
         90\% & 80\% & 28.5M & 25.1 \\
         95\% & 90\% & 14.3M & 32.2 \\
    \bottomrule
    \end{tabular}
    \vspace{.5em}
    \caption{\textbf{WikiText-103}: test perplexity for forward-backward sparsities.}
    \label{tab:wikitext}
    \end{minipage}
\end{table}
On \textbf{Enwik8}, the baseline 24-layer dense Transformer-XL obtains $0.99$ bits-per-character (BPC).
We apply \methodname to training this model and vary the forward and backward sparsity rates as shown in Figure \ref{fig:measure} (c). We find that we can obtain results comparable to the dense model all the way up to $80\%$ sparsity. When comparing to previously published models that were trained and evaluated at a modest parameter count (under 60M parameters) in Table \ref{tab:enwik8_small_models} we see that our Transformer-XL + \methodname achieves state-of-the-art performance. We also compare to magnitude pruning for a smaller Transformer model in appendix \ref{app:lm}.

On \textbf{WikiText-103} our baseline 16-layer Transformer-XL obtains $18.3$ test perplexity. When trained with \methodname, we see in Table \ref{tab:wikitext} that we can achieve 80\% sparsity with minimal performance degradation, and performance begins to drift beyond the 90\% sparsity range.  Most importantly, the sparse model is significantly better than the even the smaller dense model with $3\times$ as many parameters.

\section{Conclusion}

In this work, we considered the question of effectively and efficiently training sparse neural networks. Performant sparse networks promise to democratise research with their low-resource usage, provide savings on compute and memory and also allow the proportional scaling up of model sizes. Prior works have shown the efficacy of pruning dense neural networks to highly sparse equivalents that are able to retain most of their original performance. Motivated by these successes, more recent works have attempted to maintain fully sparse networks throughout training. While a lot of progress has been made, most of these still involve the calculation of some dense weights or gradients, or involve operations that cannot be efficiently implemented with today's tools. Building on this, we introduced a novel method, \methodname that stays fully sparse in the both the backward and forward passes and is able to be implemented easily with modern neural network packages. Our method involves keeping around only the highest weights by magnitude in the forward pass and an extra set of \textit{exploration} weights in the backward. Practitioners can choose their own values for both sparsities, based on the resource budget available. We further introduced a novel form of regularisation to encourage exploration in weight space. Coupled with this loss, \methodname achieves comparable performance to existing dense-to-sparse methods on ImageNet while remaining sparse, and exceeding the performance of several sparse-to-sparse methods. We further demonstrated the efficacy of our method on language modeling, the first such method to successfully sparsify Transformers in this context. We're also able to achieve state-of-art results for small models, with $1.00$ bpc at $55M$ parameters (versus a baseline of $0.99$ at 277M parameters). While these are encouraging findings, more work is required to fully integrate \methodname with sparse hardware and the appropriate sparse kernels. We hope practioners and researchers alike find our method useful for reducing computational requirements, and to build on for even more powerful methods of sparsification. 

\section*{Acknowledgements}

We'd like to thank Jacob Menick, Karen Simonyan,  Tim Harley and Malcolm Reynolds for their helpful feedback throughout the project. We'd also like to thank Utku Evci for their help with running baselines for the ImageNet experiments. 

\section*{Broader Impact}

Our work proposes a new method to train sparse neural networks that allows them to remain sparse throughout training -- thereby enabling a practitioner to increase the model size that can be trained on a given piece of hardware. (This would also impact deployment too, in the case of on-device or real-time learning.)
As we note in our introduction this scale-enabling should benefit the democratisation of deep learning since state-of-the-art models are ever increasing in size. 
Furthermore, there are beneficial impacts to be expected by reducing the computational footprint and energy consumption for training neural networks, as well as the higher-order impacts achieved if our work promotes the adoption of sparse networks more broadly -- thereby also reducing the deployment/inference costs.
While we do not expect any direct negative consequences from this work, the proposed method is general and widely applicable. 
We believe that the benefits offered by advances in machine learning net outweigh (by a significant margin) the potential risks and negative consequences. 
However, the technology as a whole is not purely good or benign.
As one suggestion for future research building on our contribution, we would encourage colleagues who extend or apply our work to help us assess whether the inductive biases promoted by our sparsification methods have lead to any differential sensitivity to class imbalances or other aspects of the underlying data, relative to dense counterpart approaches for a given application. Since such issues could exacerbate problems related to algorithmic bias.

\small
\bibliographystyle{plainnat}
\bibliography{references}

\begin{thebibliography}{45}
\providecommand{\natexlab}[1]{#1}
\providecommand{\url}[1]{\texttt{#1}}
\expandafter\ifx\csname urlstyle\endcsname\relax
  \providecommand{\doi}[1]{doi: #1}\else
  \providecommand{\doi}{doi: \begingroup \urlstyle{rm}\Url}\fi

\bibitem[FBw()]{FBwavernnclone}
A highly efficient real-time text-to-speech system deployed on cpus.
\newblock URL
  \url{https://ai.facebook.com/blog/a-highly-efficient-real-time-text-to-speech-system-deployed-on-cpus/}.

\bibitem[Baevski and Auli(2019)]{baevski2018adaptive}
Alexei Baevski and Michael Auli.
\newblock Adaptive input representations for neural language modeling.
\newblock In \emph{International Conference on Learning Representations}, 2019.
\newblock URL \url{https://openreview.net/forum?id=ByxZX20qFQ}.

\bibitem[Bellec et~al.(2018)Bellec, Kappel, Maass, and Legenstein]{Bellec2017}
Guillaume Bellec, David Kappel, Wolfgang Maass, and Robert~A. Legenstein.
\newblock Deep rewiring: Training very sparse deep networks.
\newblock In \emph{International Conference on Learning Representations}, 2018.

\bibitem[Brafman and Tennenholtz(2003)]{BrafmanOptimism}
Ronen~I. Brafman and Moshe Tennenholtz.
\newblock R-max - a general polynomial time algorithm for near-optimal
  reinforcement learning.
\newblock \emph{J. Mach. Learn. Res.}, 3\penalty0 (null):\penalty0 213–231,
  March 2003.
\newblock ISSN 1532-4435.
\newblock \doi{10.1162/153244303765208377}.
\newblock URL \url{https://doi.org/10.1162/153244303765208377}.

\bibitem[Christos~Louizos(2018)]{Louizos2018}
Diederik P.~Kingma Christos~Louizos, Max~Welling.
\newblock Learning sparse neural networks through $l_0$ regularization.
\newblock In \emph{International Conference on Learning Representations}, 2018.

\bibitem[Dai et~al.(2019)Dai, Yang, Yang, Carbonell, Le, and
  Salakhutdinov]{dai2019transformer}
Zihang Dai, Zhilin Yang, Yiming Yang, Jaime Carbonell, Quoc~V Le, and Ruslan
  Salakhutdinov.
\newblock Transformer-xl: Attentive language models beyond a fixed-length
  context.
\newblock \emph{arXiv preprint arXiv:1901.02860}, 2019.

\bibitem[Evci et~al.(2019{\natexlab{a}})Evci, Gale, Menick, Castro, and
  Elsen]{evci2019rigging}
Utku Evci, Trevor Gale, Jacob Menick, Pablo~Samuel Castro, and Erich Elsen.
\newblock Rigging the lottery: Making all tickets winners, 2019{\natexlab{a}}.

\bibitem[Evci et~al.(2019{\natexlab{b}})Evci, Pedregosa, Gomez, and
  Elsen]{Evci2019}
Utku Evci, Fabian Pedregosa, Aidan~N. Gomez, and Erich Elsen.
\newblock The difficulty of training sparse neural networks.
\newblock \emph{ArXiv}, 2019{\natexlab{b}}.
\newblock URL \url{http://arxiv.org/abs/1906.10732}.

\bibitem[Frankle and Carbin(2019)]{frankle2018}
Jonathan Frankle and Michael Carbin.
\newblock The lottery ticket hypothesis: Finding sparse, trainable neural
  networks.
\newblock In \emph{7th International Conference on Learning Representations,
  {ICLR} 2019, New Orleans, LA, USA, May 6-9, 2019}, 2019.
\newblock URL \url{https://openreview.net/forum?id=rJl-b3RcF7}.

\bibitem[Gale et~al.(2019)Gale, Elsen, and Hooker]{gale2019state}
Trevor Gale, Erich Elsen, and Sara Hooker.
\newblock The state of sparsity in deep neural networks.
\newblock \emph{CoRR}, abs/1902.09574, 2019.
\newblock URL \url{http://arxiv.org/abs/1902.09574}.

\bibitem[García-Martín et~al.(2019)García-Martín, Rodrigues, Riley, and
  Grahn]{GARCIAMARTIN201975}
Eva García-Martín, Crefeda~Faviola Rodrigues, Graham Riley, and Håkan Grahn.
\newblock Estimation of energy consumption in machine learning.
\newblock \emph{Journal of Parallel and Distributed Computing}, 134:\penalty0
  75 -- 88, 2019.
\newblock ISSN 0743-7315.
\newblock \doi{https://doi.org/10.1016/j.jpdc.2019.07.007}.
\newblock URL
  \url{http://www.sciencedirect.com/science/article/pii/S0743731518308773}.

\bibitem[Graves(2013)]{graves2013generating}
Alex Graves.
\newblock Generating sequences with recurrent neural networks.
\newblock \emph{arXiv preprint arXiv:1308.0850}, 2013.

\bibitem[Ha et~al.(2016)Ha, Dai, and Le]{ha2016hypernetworks}
David Ha, Andrew Dai, and Quoc~V Le.
\newblock Hypernetworks.
\newblock \emph{arXiv preprint arXiv:1609.09106}, 2016.

\bibitem[Han et~al.(2015)Han, Pool, Tran, and Dally]{han2015learning}
Song Han, Jeff Pool, John Tran, and William Dally.
\newblock Learning both weights and connections for efficient neural network.
\newblock In \emph{Advances in neural information processing systems}, 2015.

\bibitem[{He} et~al.(2019){He}, {Sainath}, {Prabhavalkar}, {McGraw}, {Alvarez},
  {Zhao}, {Rybach}, {Kannan}, {Wu}, {Pang}, {Liang}, {Bhatia}, {Shangguan},
  {Li}, {Pundak}, {Sim}, {Bagby}, {Chang}, {Rao}, and
  {Gruenstein}]{googleondeviceasr}
Y.~{He}, T.~N. {Sainath}, R.~{Prabhavalkar}, I.~{McGraw}, R.~{Alvarez},
  D.~{Zhao}, D.~{Rybach}, A.~{Kannan}, Y.~{Wu}, R.~{Pang}, Q.~{Liang},
  D.~{Bhatia}, Y.~{Shangguan}, B.~{Li}, G.~{Pundak}, K.~C. {Sim}, T.~{Bagby},
  S.~{Chang}, K.~{Rao}, and A.~{Gruenstein}.
\newblock Streaming end-to-end speech recognition for mobile devices.
\newblock In \emph{ICASSP 2019 - 2019 IEEE International Conference on
  Acoustics, Speech and Signal Processing (ICASSP)}, pages 6381--6385, 2019.

\bibitem[Jouppi et~al.(2017)Jouppi, Young, Patil, Patterson, Agrawal, Bajwa,
  Bates, Bhatia, Boden, Borchers, et~al.]{jouppi2017datacenter}
Norman~P Jouppi, Cliff Young, Nishant Patil, David Patterson, Gaurav Agrawal,
  Raminder Bajwa, Sarah Bates, Suresh Bhatia, Nan Boden, Al~Borchers, et~al.
\newblock In-datacenter performance analysis of a tensor processing unit.
\newblock In \emph{Proceedings of the 44th Annual International Symposium on
  Computer Architecture}, pages 1--12, 2017.

\bibitem[Kalchbrenner et~al.(2018)Kalchbrenner, Elsen, Simonyan, Noury,
  Casagrande, Lockhart, Stimberg, Oord, Dieleman, and
  Kavukcuoglu]{kalchbrenner2018}
Nal Kalchbrenner, Erich Elsen, Karen Simonyan, Seb Noury, Norman Casagrande,
  Edward Lockhart, Florian Stimberg, Aaron Oord, Sander Dieleman, and Koray
  Kavukcuoglu.
\newblock Efficient neural audio synthesis.
\newblock In \emph{International Conference on Machine Learning (ICML)}, 2018.

\bibitem[Kaplan et~al.(2020)Kaplan, McCandlish, Henighan, Brown, Chess, Child,
  Gray, Radford, Wu, and Amodei]{kaplan2020scaling}
Jared Kaplan, Sam McCandlish, Tom Henighan, Tom~B. Brown, Benjamin Chess, Rewon
  Child, Scott Gray, Alec Radford, Jeffrey Wu, and Dario Amodei.
\newblock Scaling laws for neural language models, 2020.

\bibitem[Krause et~al.(2016)Krause, Lu, Murray, and
  Renals]{krause2016multiplicative}
Ben Krause, Liang Lu, Iain Murray, and Steve Renals.
\newblock Multiplicative lstm for sequence modelling.
\newblock \emph{arXiv preprint arXiv:1609.07959}, 2016.

\bibitem[Kusupati et~al.(2020)Kusupati, Ramanujan, Somani, Wortsman, Jain,
  Kakade, and Farhadi]{kusupati2020soft}
Aditya Kusupati, Vivek Ramanujan, Raghav Somani, Mitchell Wortsman, Prateek
  Jain, Sham Kakade, and Ali Farhadi.
\newblock Soft threshold weight reparameterization for learnable sparsity,
  2020.

\bibitem[LeCun et~al.(1990)LeCun, Denker, and Solla]{lecun1990}
Yann LeCun, John~S. Denker, and Sara~A. Solla.
\newblock {Optimal Brain Damage}.
\newblock In \emph{Advances in Neural Information Processing Systems}, 1990.

\bibitem[Lenc et~al.(2019)Lenc, Elsen, Schaul, and Simonyan]{karelhybrides}
Karel Lenc, Erich Elsen, Tom Schaul, and Karen Simonyan.
\newblock Non-differentiable supervised learning with evolution strategies and
  hybrid methods.
\newblock \emph{CoRR}, abs/1906.03139, 2019.
\newblock URL \url{http://arxiv.org/abs/1906.03139}.

\bibitem[{Lin} et~al.(2017){Lin}, {Yu}, {Zhang}, {Yang}, {Zhang}, and
  {Zhao}]{edgeprivacy2}
J.~{Lin}, W.~{Yu}, N.~{Zhang}, X.~{Yang}, H.~{Zhang}, and W.~{Zhao}.
\newblock A survey on internet of things: Architecture, enabling technologies,
  security and privacy, and applications.
\newblock \emph{IEEE Internet of Things Journal}, 4\penalty0 (5):\penalty0
  1125--1142, 2017.

\bibitem[Mahoney(2011)]{mahoney2011large}
Matt Mahoney.
\newblock Large text compression benchmark.
\newblock \emph{URL: http://www. mattmahoney. net/text/text. html}, 2011.

\bibitem[Merity et~al.(2016)Merity, Xiong, Bradbury, and
  Socher]{merity2016pointer}
Stephen Merity, Caiming Xiong, James Bradbury, and Richard Socher.
\newblock Pointer sentinel mixture models.
\newblock \emph{arXiv preprint arXiv:1609.07843}, 2016.

\bibitem[Mocanu et~al.(2018)Mocanu, Mocanu, Stone, Nguyen, Gibescu, and
  Liotta]{Mocanu2018}
Decebal~Constantin Mocanu, Elena Mocanu, Peter Stone, Phuong~H Nguyen,
  Madeleine Gibescu, and Antonio Liotta.
\newblock \href{http://www.nature.com/articles/s41467-018-04316-3}{Scalable
  training of artificial neural networks with adaptive sparse connectivity
  inspired by network science}.
\newblock \emph{Nature Communications}, 2018.

\bibitem[Molchanov et~al.(2017)Molchanov, Ashukha, and
  Vetrov]{variational-dropout}
Dmitry Molchanov, Arsenii Ashukha, and Dmitry~P. Vetrov.
\newblock Variational {D}ropout {S}parsifies {D}eep {N}eural {N}etworks.
\newblock In \emph{Proceedings of the 34th International Conference on Machine
  Learning, {ICML} 2017, Sydney, NSW, Australia, 6-11 August 2017}, pages
  2498--2507, 2017.

\bibitem[Mostafa and Wang(2019)]{Mostafa2019}
Hesham Mostafa and Xin Wang.
\newblock Parameter efficient training of deep convolutional neural networks by
  dynamic sparse reparameterization.
\newblock In \emph{Proceedings of the 36th International Conference on Machine
  Learning, {ICML} 2019, 9-15 June 2019, Long Beach, California, {USA}}, pages
  4646--4655, 2019.
\newblock URL \url{http://proceedings.mlr.press/v97/mostafa19a.html}.

\bibitem[Narang et~al.(2017)Narang, Diamos, Sengupta, and
  Elsen]{exploring-sparsity-rnn}
Sharan Narang, Greg Diamos, Shubho Sengupta, and Erich Elsen.
\newblock Exploring sparsity in recurrent neural networks.
\newblock In \emph{5th International Conference on Learning Representations,
  {ICLR} 2017, Toulon, France, April 24-26, 2017, Conference Track
  Proceedings}, 2017.
\newblock URL \url{https://openreview.net/forum?id=BylSPv9gx}.

\bibitem[NVIDIA(2020)]{nvidiaampere}
NVIDIA.
\newblock Nvidia a100 tensor core gpu architecture, 2020.
\newblock URL
  \url{https://www.nvidia.com/content/dam/en-zz/Solutions/Data-Center/nvidia-ampere-architecture-whitepaper.pdf}.

\bibitem[Pang et~al.(2018)Pang, Sainath, Prabhavalkar, Gupta, Wu, Zhang, and
  Chiu]{Pang2018}
Ruoming Pang, Tara Sainath, Rohit Prabhavalkar, Suyog Gupta, Yonghui Wu,
  Shuyuan Zhang, and Chung-Cheng Chiu.
\newblock Compression of end-to-end models.
\newblock In \emph{Proc. Interspeech 2018}, pages 27--31, 2018.
\newblock \doi{10.21437/Interspeech.2018-1025}.
\newblock URL \url{http://dx.doi.org/10.21437/Interspeech.2018-1025}.

\bibitem[Radford et~al.(2019)Radford, Wu, Child, Luan, Amodei, and
  Sutskever]{radford2019language}
Alec Radford, Jeffrey Wu, Rewon Child, David Luan, Dario Amodei, and Ilya
  Sutskever.
\newblock Language models are unsupervised multitask learners.
\newblock \emph{OpenAI Blog}, 1\penalty0 (8):\penalty0 9, 2019.

\bibitem[{Schwartz} et~al.(2019){Schwartz}, {Dodge}, {Smith}, and
  {Etzioni}]{greenAI}
Roy {Schwartz}, Jesse {Dodge}, Noah~A. {Smith}, and Oren {Etzioni}.
\newblock {Green AI}.
\newblock \emph{arXiv e-prints}, art. arXiv:1907.10597, Jul 2019.

\bibitem[Shoeybi et~al.(2019)Shoeybi, Patwary, Puri, LeGresley, Casper, and
  Catanzaro]{shoeybi2019megatronlm}
Mohammad Shoeybi, Mostofa Patwary, Raul Puri, Patrick LeGresley, Jared Casper,
  and Bryan Catanzaro.
\newblock Megatron-lm: Training multi-billion parameter language models using
  model parallelism, 2019.

\bibitem[Stanley and Miikkulainen(2002)]{NEAT}
Kenneth~O. Stanley and Risto Miikkulainen.
\newblock Evolving neural networks through augmenting topologies.
\newblock \emph{Evol. Comput.}, 10\penalty0 (2):\penalty0 99–127, June 2002.
\newblock ISSN 1063-6560.
\newblock \doi{10.1162/106365602320169811}.
\newblock URL \url{https://doi.org/10.1162/106365602320169811}.

\bibitem[Str\"om(1997)]{sparse-connection-1997}
Nikko Str\"om.
\newblock Sparse {C}onnection and {P}runing in {L}arge {D}ynamic {A}rtificial
  {N}eural {N}etworks.
\newblock In \emph{EUROSPEECH}, 1997.

\bibitem[Strubell et~al.(2019)Strubell, Ganesh, and
  McCallum]{Strubell2019EnergyAP}
Emma Strubell, Ananya Ganesh, and Andrew McCallum.
\newblock Energy and policy considerations for deep learning in nlp.
\newblock In \emph{ACL}, 2019.

\bibitem[Sukhbaatar et~al.(2019)Sukhbaatar, Grave, Lample, Jegou, and
  Joulin]{sukhbaatar2019augmenting}
Sainbayar Sukhbaatar, Edouard Grave, Guillaume Lample, Herve Jegou, and Armand
  Joulin.
\newblock Augmenting self-attention with persistent memory.
\newblock \emph{arXiv preprint arXiv:1907.01470}, 2019.

\bibitem[Tan and Le(2019)]{efficientnet}
Mingxing Tan and Quoc Le.
\newblock {E}fficient{N}et: Rethinking model scaling for convolutional neural
  networks.
\newblock 97:\penalty0 6105--6114, 09--15 Jun 2019.
\newblock URL \url{http://proceedings.mlr.press/v97/tan19a.html}.

\bibitem[Thimm and Fiesler(1995)]{Thimm95evaluatingpruning}
Georg Thimm and Emile Fiesler.
\newblock Evaluating pruning methods.
\newblock In \emph{National Chiao-Tung University}, page~2, 1995.

\bibitem[Vaswani et~al.(2017)Vaswani, Shazeer, Parmar, Uszkoreit, Jones, Gomez,
  Kaiser, and Polosukhin]{vaswani2017attention}
Ashish Vaswani, Noam Shazeer, Niki Parmar, Jakob Uszkoreit, Llion Jones,
  Aidan~N Gomez, {\L}ukasz Kaiser, and Illia Polosukhin.
\newblock Attention is all you need.
\newblock In \emph{Advances in neural information processing systems}, pages
  5998--6008, 2017.

\bibitem[Wortsman et~al.(2019)Wortsman, Farhadi, and Rastegari]{DNW}
Mitchell Wortsman, Ali Farhadi, and Mohammad Rastegari.
\newblock Discovering neural wirings.
\newblock In H.~Wallach, H.~Larochelle, A.~Beygelzimer, F.~d\'Alch\'{e} Buc,
  E.~Fox, and R.~Garnett, editors, \emph{Advances in Neural Information
  Processing Systems 32}, pages 2684--2694. Curran Associates, Inc., 2019.
\newblock URL
  \url{http://papers.nips.cc/paper/8536-discovering-neural-wirings.pdf}.

\bibitem[{Zhang} et~al.(2018){Zhang}, {Chen}, {Zhao}, {Cheng}, and
  {Hu}]{edgeprivacy1}
J.~{Zhang}, B.~{Chen}, Y.~{Zhao}, X.~{Cheng}, and F.~{Hu}.
\newblock Data security and privacy-preserving in edge computing paradigm:
  Survey and open issues.
\newblock \emph{IEEE Access}, 6:\penalty0 18209--18237, 2018.

\bibitem[Zhou et~al.(2019)Zhou, Lan, Liu, and Yosinski]{deconstructing_lottery}
Hattie Zhou, Janice Lan, Rosanne Liu, and Jason Yosinski.
\newblock \href{http://arxiv.org/abs/1905.01067}{Deconstructing Lottery
  Tickets: Zeros, Signs, and the Supermask}.
\newblock \emph{ArXiv}, 2019.

\bibitem[Zhu and Gupta(2018)]{gupta2018}
Michael Zhu and Suyog Gupta.
\newblock \href{https://arxiv.org/abs/1710.01878}{To Prune, or Not to Prune:
  Exploring the Efficacy of Pruning for Model Compression}.
\newblock In \emph{International Conference on Learning Representations
  Workshop}, 2018.

\end{thebibliography}

\newpage
\appendix

\section*{Supplementary}
\section{Language Modeling}
\label{app:lm}
We train our Transformer-XL models with a very similar setup to \citet{dai2019transformer}. The dense model hyper-parameters are listed in Table \ref{tab:lm_hp}. We train with a learning rate warmup for 4000 steps from 1e-7 up to a value of 2e-4 and then apply a cosine decay. For WikiText-103 and enwik8 our dense model uses the same specification as the large Transformer-XL in \citet{dai2019transformer}, which has 285M parameters\footnote{The original publication erroneously listed 255M parameters, however it has been clarified as 285M with the authors.}. 

\begin{table}[h]
    \centering
    \begin{tabular}{lcc}
   \toprule
   & \textbf{Enwik8} & \textbf{WikiText-103} \\
    \midrule
    \texttt{num layers} & 24 & 18 \\
    \texttt{$d_{model}$} & 1024 & 1024 \\
    \texttt{$d_{ff}$} & 3072 & 4096 \\
    \texttt{$d_{embed}$} & 512 & adaptive: \citep{baevski2018adaptive} \\
    \texttt{tie input/output embeddings} & true & tie head only: \citep{baevski2018adaptive} \\
    \texttt{num heads} & 8 & 16 \\
    \texttt{dropout} & 0.05-0.2 & 0.05 - 0.2 \\
    \texttt{learning rate} & 2e-4 & 2e-4 \\
    \texttt{grad clip by global norm} & 0.25 & 0.25 \\
    \texttt{window size} & 768 & 512 \\
    \texttt{train mem size} & 2304 & 768 \\
    \texttt{eval mem size} & 5000 & 2000 \\
    \texttt{num params} & 69M & 285M \\
    \bottomrule
    \end{tabular}
    \vspace{0.5em}
    \caption{Transformer-XL baseline hyper-parameters.}
    \label{tab:lm_hp}
\end{table}

We further compare to a magnitude pruning baseline on enwik8. We found we were unable to implement this with the large model due to the additional memory requirements. Instead we compare \methodname and pruning on a smaller version of the Transformer-XL model of $69M$ parameters. This has identical training and hyper-parameters to below with the exception of $d_{model}=512$, $d_{ff}=1536$ and $num_heads=8$. We summarise the results below. We find that pruning slightly outperforms \methodname when \methodname is allowed a dense backward (albeit the forward pass is also sparse). However, \methodname is competitive even in the regime of sparse backward passes.

\begin{table}[h]
    \centering
  
  \begin{tabular}{c c c c c}
    \toprule
        \textbf{Fwd} & \textbf{Bwd} & \textbf{Params} & \textbf{Pruning BPC} & \textbf{\methodname BPC} \\
    \midrule
         0\% & 0\% & 69M & 1.00 & 1.00 \\
         \midrule
         80\% & 0\% & 14M & 1.02 & 1.03 \\
         80\% & 60\% & 14M & - & 1.05\\
         90\% & 0\%  & 7M & 1.06 &1.08 \\
         90\% & 80\% & 7M & - & 1.10 \\
         95\% & 0\% & 1.4M & 1.13 & 1.14 \\
         95\% & 90\% & 1.4M & - & 1.17 \\

    \bottomrule
    \end{tabular}
    \vspace{.5em}
    \caption{\textbf{enwik8}: test perplexity for the smaller transformer model.}
\end{table}

\section{ImageNet}
\label{app:imagenet}

For all ImageNet experiments we use a ResNet-50 set up as in prior work \cite{gale2019state}. We use a batch size of $4096$ and train for $32000$ steps. We use use a learning rate of $1.6$ (with a linear ramp up for $5$ epochs) followed by learning rate drops by factors of $0.1$ at $30$, $70$ and $90$ epochs. For \methodname we use a weight decay of $1e-4$, and train for a range of backward and forward sparsity rates. 

For our experiments we keep the first and last layers dense as in previous works \cite{gale2019state, evci2019rigging}. We also relax the assumption and show below the performance if all layers are sparsified. 
\begin{figure}
\centering
\includegraphics[width=0.4\textwidth]{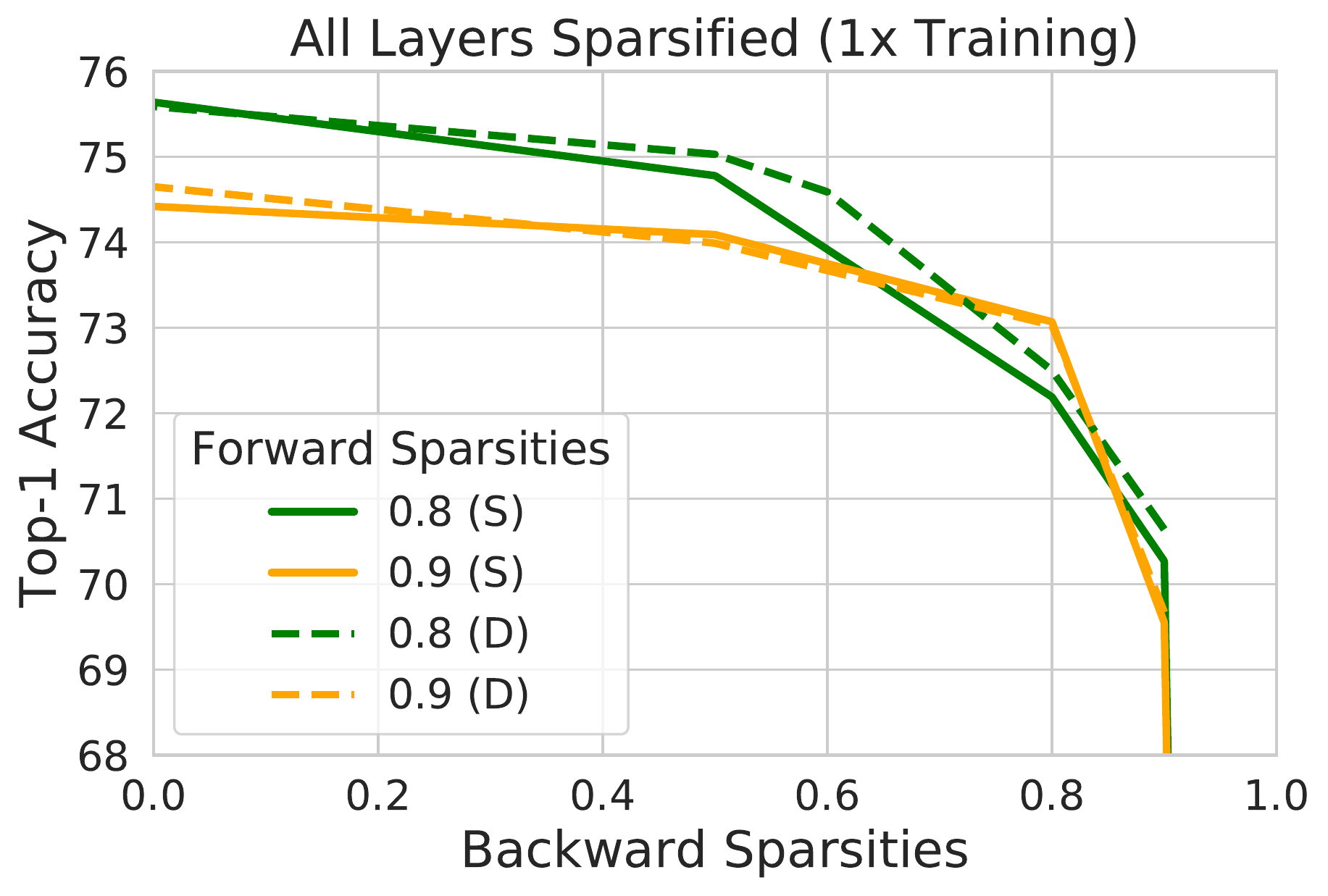}
\end{figure}

\section{Implementation of RigL and \methodname}
\label{app:impl}

In the sections above we compared briefly the implementations of RigL and \methodname and argued the relative ease of implementing \methodname because of some of the practical constraints a theoretically sparse implementation of RigL faces.

We first detail how RigL might actually be implemented and the difficulties that would be encountered.  RigL occasionally requires calculating the \texttt{Top-K} values and locations of the full dense gradient with respect the parameters for every layer.  The usual framework encapsulation is that all the gradients are computed and then sent to the optimiser.  Doing the \texttt{Top-K} in the optimiser has the advantage of not needing modify the gradient calculations, but the large downside of meaning that the dense gradient would need to be materialised.  This means the \texttt{Top-K} must happen inside the gradient calculation. 

The type returned by the gradient calculation must be consistent, so it must always return both gradient values and locations and it must accept as arguments locations and a step count.  If the step count indicates a \texttt{Top-K} over a dense gradient is to be performed, then input locations are ignored and the output locations contain updated locations.  Otherwise, the input locations are used and simply copied to the output.

Inside the actual gradient calculation, it must `chunk' the calculation of the dense gradient so as maintain a bound on the memory required.  Assuming a data parallel regime, after each chunk is calculated locally, it must then be all-reduced.  Then on each replica the running \texttt{Top-K} values are concatenated with the gradient chunk and a new running \texttt{Top-K} is calculated from this list.  This process must proceed completely serially to maintain the memory bounds.

The serialisation introduces some perhaps non-trivial overheads, but most problematic is that no gradient calculations currently work like this.  Every gradient calculation would need to re-written to do the appropriate chunking, this is both a high burden as this code involve rewriting a great deal of code.  And it also introduces its own performance ramifications.  Common libraries and/or data formats, especially for convolutions, might not support strides that would be necessary to compute arbitrary output shapes.  If they do, it might come with negative performance implications.

Lastly, we show results for an implementation of \methodname that only requires calculating the \texttt{Top-K} every $N$ steps, where $N=100$ (as opposed to $N=1$, which corresponds to performing this every iteration). Such an implementation only requires occasional communication of the indices and weights and the \texttt{Top-K} operation can be calculated in parallel on CPU as it does not require any data or forward passes. The accelerator need only know the actual sparse weights and can be implemented entirely sparsely. We run \methodname for a variety of sparsity fractions and report the results below:

\begin{table}[h]
    \centering
  
  \begin{tabular}{c c  c c}
    \toprule
        \textbf{Fwd} & \textbf{Bwd}  & \textbf{$N=1$} & \textbf{$N=100$} \\
    \midrule
         80\% & 50\% & 75.03 & 75.14 \\
         90\% & 80\% & 73.03 & 73.18 \\
         95\% & 90\% & 70.42 & 70.38 \\

    \bottomrule
    \end{tabular}
    \vspace{.5em}
    \caption{\methodname at different frequencies of \texttt{Top-K}}
\end{table}

\section{Pseudocode}

In general \methodname can be implemented by modifying the parameters used in the forward pass and applying a gradient with respect to only some of the weighs in the backward pass. Below we demonstrate how this could be implemented with existing dense kernels and explicit masking of the weights. For a truly sparse implementation, custom sparse kernels would be required. 

\begin{algorithm}[H]
\caption{TopKAST}
// First perform a Top-K \\
dense\_params = initialise() \\
fwd\_params = TopK(dense\_params, X\%) \\
bwd\_params = TopK(dense\_params, Y\%) \\
just\_bwd\_set = set(bwd\_params) - set(fwd\_params) \\
...  \\
// Output with just the TopK params \\
output = model(fwd\_params, input) \\
loss = loss\_fn(output) \\ \\
// Exploration L2 Loss \\ 
loss += l2(fwd\_params) + l2(just\_bwd\_set) / (X/100) \\
... \\
// Update only the bwd params \\
bwd\_params = bwd\_params - grad(loss, bwd\_params)
\end{algorithm}

\end{document}